\documentclass[10pt,twocolumn,letterpaper]{article}
\usepackage[pagenumbers]{cvpr}
\usepackage{graphicx}
\usepackage{booktabs}
\usepackage{multirow}
\usepackage{amsmath,amssymb,amsthm}
\newtheorem{proposition}{Proposition}
\usepackage{microtype}
\usepackage{enumitem}
\usepackage{pifont}
\usepackage{tabularx}
\usepackage{siunitx}

\usepackage{colortbl}
\definecolor{resultshade}{RGB}{239,244,247}

\usepackage{float}
\usepackage{algorithm}
\usepackage{algpseudocode}
\newtheorem{suppthm}{Theorem}

\newtheorem{supplemma}[suppthm]{Lemma}
\newtheorem{suppcor}[suppthm]{Corollary}
\newtheorem{suppprop}[suppthm]{Proposition}
\theoremstyle{definition}
\newtheorem{suppdef}[suppthm]{Definition}
\theoremstyle{plain}

\definecolor{cvprblue}{rgb}{0.21,0.49,0.74}
\usepackage[pagebackref,breaklinks,colorlinks,allcolors=cvprblue]{hyperref}
\usepackage[capitalize]{cleveref}

\def\confName{CVPR}
\def\confYear{2027}

\title{ViStep-RL: Stepwise Reinforcement Learning for Long-Horizon Visual Agents}
\author{%
\parbox{\dimexpr\textwidth-2\tabcolsep\relax}{\centering
\normalsize\bfseries
Junyang Chen$^{1,2}$,\quad Yao Qin$^{3}$,\quad Xiao Li$^{1,2}$,\quad Ningchao Wang$^{4}$\\[3pt]
Yiguo Qiao$^{1,2,*}$,\quad Wenbin Li$^{4}$,\quad Xin Geng$^{1,2}$\\[7pt]
\small\normalfont
$^{1}$School of Computer Science and Engineering, Southeast University, Nanjing 211189, China\\[2pt]
$^{2}$Key Laboratory of Computer Network and Information Integration (Ministry of Education),\\
Southeast University, Nanjing 211189, China\\[2pt]
$^{3}$University of Science and Technology of China, Hefei, China\\[2pt]
$^{4}$University of Bath, Bath, United Kingdom\\[3pt]
{\footnotesize $^{*}$Corresponding author.}
}}

\begin{document}
\maketitle
\begin{abstract}
    Long-horizon visual agents must do more than inspect images: they must turn observations into evidence that remains usable as a task unfolds. Existing active-perception methods teach agents where and how to look, while a final-answer reward gives little structure for deciding which observations satisfy dependent subgoals or should be retained for later use. We present \textbf{ViStep-RL} (\textbf{Vi}sual \textbf{Step}wise Reinforcement Learning), which models execution as a dependency-linked graph of evidence-bearing subgoals. A verifier tracks which subgoals are supported by valid evidence, and changes in graph potential provide a net-progress learning signal; an active memory protocol preserves evidence across turns and links retrieval to downstream use. On \textbf{ViInteract}, ViStep-RL reaches 60.0\% strict completion with Qwen2.5-VL-7B, compared with 10.0\% for the base model and 46.5\% after supervised fine-tuning. Controlled studies show that incremental task-state rewards outperform cumulative rewards by 3.2 points, while active evidence writing improves strict completion by 7.75 points over automatic writing. Under a common interaction framework, the task-trained ViStep-RL policy achieves 59.5\% answer accuracy, 24.0 points above the strongest of six general-purpose reference models. Held-out episodes confirm the gain at both Qwen scales, and gains extend to three external visual reasoning benchmarks.
\end{abstract}

\section{Introduction}
\label{sec:intro}

Long-horizon visual tasks require agents to integrate information across a sequence of observations, tool calls, and decisions. An agent may inspect a frame, identify an object, compare it with an earlier image, and rely on that comparison several steps later. Beyond deciding where to look, the agent must convert observations into evidence that remains available to support dependent task steps. When a useful observation is forgotten, detached from its source, or used before its prerequisites are established, subsequent reasoning may fail to complete the task.

Recent reinforcement-learning methods make visual perception more active. ViGoRL grounds reasoning in image regions and learns when to zoom; MGPO extends grounding to multi-turn high-resolution inspection; DeepEyes encourages image operations during reasoning; and ACTIVE-o3 trains active-perception policies~\cite{sarch2025vigorl,huang2025mgpo,zheng2025deepeyes,zhu2025activeo3}. These methods establish that an agent can learn how to acquire visual information. Long-horizon execution additionally requires tracking when acquired information qualifies as task evidence, which subsequent decisions depend on it, and whether it remains accessible as the visual context changes. Treating observations as isolated perceptual outputs leaves this evolving evidence state implicit.

We make the evidence state explicit by representing each task as a dependency graph. Each node encodes a verifiable subgoal and its evidence requirements. A node is verified only when both its evidence requirements and prerequisites are satisfied. This representation allows the verifier to track established subgoals, unresolved dependencies, and evidence that must remain available for subsequent execution. It also distinguishes genuine task progress from repeated interaction that leaves the underlying state unchanged.

\textbf{ViStep-RL} (\textbf{Vi}sual \textbf{Step}wise Reinforcement Learning) uses this evolving task state to guide agent training. After each transition, a verifier evaluates the evidence and checks whether subgoal dependencies are satisfied. Changes in normalized graph potential provide a signed progress signal for increases and decreases in task progress. An active memory protocol records provenance-linked observations, retrieves them after changes in image context, and connects retrieval to the dependent subgoal that uses the evidence. The task graph and expected evidence remain outside the policy context: during training, the verifier supplies the transition rewards used to compare actions at similar stages of the task. This design links visual acquisition, evidence persistence, and downstream task completion in one execution objective.

We evaluate ViStep-RL on \textbf{ViInteract}, a benchmark comprising complete visual-agent episodes spanning temporal interaction, multi-image reasoning, visual-tool use, OCR, pose, and mask/measurement tasks. Its primary evaluation set, Core200, contains 200 episodes, with the domain composition reported in the supplement. Beyond region localization and observation requests, ViInteract evaluates whether an agent completes the task through a sequence of visual actions, tool calls, and state updates. On Core200, ViStep-RL improves Strict completion over SFT by 13.5 and 5.5 percentage points for Qwen2.5-VL-7B and Qwen2.5-VL-3B, respectively. Under matched training and interaction budgets, ViStep-RL also improves Strict over standard terminal-answer GRPO by 11.5 points at 7B and 5.0 points at 3B. On the separate Process500 panel, incremental graph rewards improve Strict completion by 3.2 percentage points over cumulative graph rewards. In the writing--retrieval factorial, active evidence writing improves Strict completion by 7.75 percentage points over automatic writing, averaged across retrieval conditions. With retrieval held fixed, active writing also improves final-answer accuracy by 8.0 points with learned retrieval and 7.0 points with BM25. Together, these results support the complementary roles of progress-based training and evidence persistence in completing visual tasks. Evaluations of task-trained variants on MMLongBench-Doc, MuirBench, and MMSI-Bench further examine document, multi-image, and spatial reasoning. Our main contributions are summarized as follows:

\begin{itemize}[leftmargin=*,itemsep=2pt,topsep=3pt]
    \item We propose \textbf{ViStep-RL}, a verifier-guided framework for long-horizon visual execution. Its dependency-aware graph reward uses changes in task-state potential to credit progress without repeatedly rewarding unchanged states.
    
    \item A task-coupled memory mechanism integrates evidence writing and retrieval with downstream verification, preserving provenance-linked observations across turns.
    
    \item We introduce \textbf{ViInteract}, a benchmark for end-to-end visual-agent execution, and evaluate our approach across diverse tasks and two model scales.

\end{itemize}

\section{Related Work}
\label{sec:related}

\paragraph{Visual reasoning, grounding, and benchmarks.}
Vision-language research has evolved from visual question answering to grounding, high-resolution perception, and multi-image reasoning. Early benchmarks evaluate visual and textual evidence~\cite{goyal2017vqa,hudson2019gqa,singh2019textvqa,mathew2021docvqa,masry2022chartqa}, while recent suites extend to compositional perception and multimodal reasoning~\cite{fu2023mme,yu2023mmvet,yue2024mmmu,lu2024mathvista,fu2024blink,ying2024mmt,jiang2024mantis,wang2024muirbench}. Advances in LLaVA, Qwen2-VL, and InternVL strengthen these capabilities~\cite{liu2023llava,wang2024qwen2vl,chen2024internvl}. Recent interactive approaches, including ViGoRL, MGPO, DeepEyes, and ACTIVE-o3, enable active visual grounding and perception through image operations~\cite{sarch2025vigorl,huang2025mgpo,zheng2025deepeyes,zhu2025activeo3}. Unlike these perception-centered settings, ViInteract evaluates end-to-end task completion across heterogeneous tools, dependencies, and persistent state.

\paragraph{Interactive agents and reinforcement learning.}
Tool-using agents integrate language reasoning with external actions, from ReAct and Toolformer to multimodal interactive systems~\cite{yao2023react,schick2023toolformer,wu2023visualchatgpt,yang2023mmreact}. Benchmarks such as AgentBench, WebArena, OSWorld, and Android-in-the-Wild evaluate long-horizon interaction and dynamically evolving environment states~\cite{liu2023agentbench,zhou2024webarena,xie2024osworld,rawles2023aitw}. Reinforcement learning optimizes such trajectories through policy gradients and stabilized updates, including PPO and group-relative optimization methods~\cite{williams1992reinforce,schulman2017ppo,sha2024deepseekmath,ahmadian2024rloo,yu2025dapo}. Terminal correctness alone provides sparse and delayed feedback for complex multi-step tasks. Process supervision evaluates intermediate reasoning steps~\cite{lightman2024verify}, while AgentPRM learns process rewards from Monte Carlo rollout targets~\cite{choudhury2025agentprm}. Verifiable Process Rewards use task oracles to score intermediate actions~\cite{yuan2026vpr}; PerceptRL rewards spatial fidelity during visual inspection~\cite{guan2026perceptrl}. ViStep-RL tracks what happens to an observation after inspection: its source must remain identifiable and its evidence available when a later subgoal calls for it.

\paragraph{Memory and long-horizon evidence management.}
Long-horizon agents preserve evidence through retrieval and persistent memory. RAG and RETRO introduce external retrieval~\cite{lewis2020rag,borgeaud2022retro}, while MemGPT, MemoryBank, Reflexion, Voyager, and Generative Agents explore memory and experience reuse~\cite{packer2023memgpt,zhong2023memorybank,shinn2023reflexion,wang2023voyager,park2023generativeagents}. Self-RAG and HippoRAG further investigate selective retrieval and structured memory~\cite{asai2024selfrag,gutierrez2024hipporag}. Agent Workflow Memory induces reusable action routines~\cite{wang2025awm}, and A-MEM dynamically links and updates stored memory notes~\cite{xu2025amem}. ViStep-RL focuses on whether visual evidence acquired within an episode remains supported and satisfies downstream dependencies. It incorporates memory writing and retrieval into an executable task graph, using factorial comparisons and memory-destruction interventions to assess their contribution to task progress.

\section{Method}
\label{sec:method}

\begin{figure*}[t]
\centering
\includegraphics[width=\textwidth]{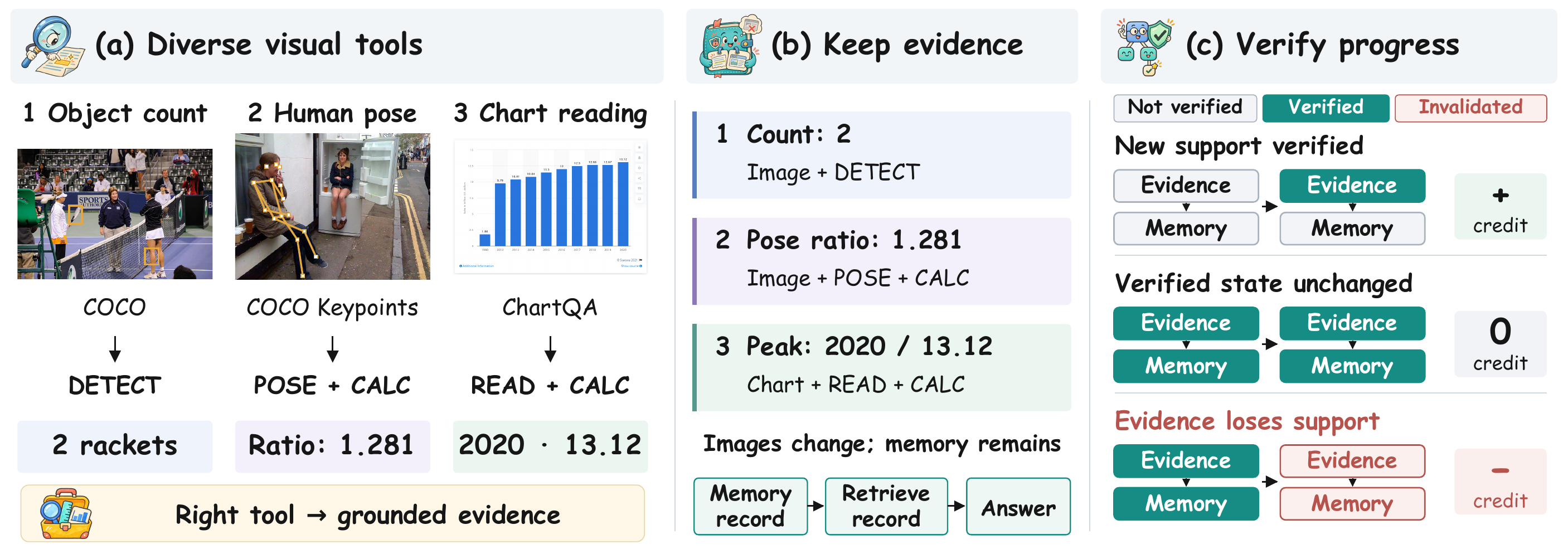}
\caption{\textbf{ViStep-RL: from visual observations to verified task progress.} (a) Source-grounded examples illustrate object counting, pose measurement, and chart reading. (b) Tool observations are stored with provenance and retrieved after visual context changes. (c) Schematic changes in strict verification: new evidence establishes progress, unchanged verification earns no additional graph credit, and invalidation retracts dependent memory verification. Auxiliary retrieval coverage is defined separately in the training potential.}
\label{fig:method-overview}
\end{figure*}

\subsection{Visual Execution and Evidence Verification}
A visual observation can be several actions old by the time the agent needs it. In a TextOCR episode, for example, the agent reads text on successive images and later identifies the image containing the requested word. Once the image leaves view, the word and its source must still be available together. Figure~\ref{fig:method-overview} shows how ViStep-RL connects these decisions. The agent chooses what to inspect, store, and retrieve. After each action, a task program verifies the evidence against explicit task requirements. The policy learns by comparing actions taken at similar stages of the same task. Each ingredient is established: verifiable task graphs, potential-based shaping~\cite{ng1999shaping} and group-relative updates. ViStep-RL ties them to one object, the provenance-tracked evidence record, which the verifier checks, the potential rewards and the action weight credits.

Formally, an episode is a finite-horizon interaction $\tau=(x_0,a_1,x_1,\ldots,a_T,x_T)$, where $x_t$ is the environment state after transition $t$. The policy samples $a_t\sim\pi_\theta(\cdot\mid h_{t-1})$ from its public history, which includes the instruction, current visual input, legal actions, and returned observations, including failures. The verifier has access to the task graph and expected evidence.

\paragraph{Execution runtime and Harness.}
The runtime executes the policy's actions within a fixed budget. It dispatches visual tools, advances the current image, handles memory operations, and records the final answer. Native evaluations use the trained agent's action protocol. For comparisons across models, the shared Harness provides a common public interface for image progression, tool calls, returned observations, and answer submission.

\paragraph{Active evidence memory.}
In native Qwen execution, advancing to a new visual turn starts a fresh prompt. The previous image, its dialogue, and the text returned by OCR leave the prompt together. Explicit memory records persist in the environment, and a search or retrieval can return their contents to the policy. The native policy can preserve a tool observation by writing its payload, source image or semantic identity, and references to the tool events that produced it. A later search can then recover both the observation and its origin. In the OCR example, this association lets the agent trace a retrieved word back to the image it must identify.

We represent the required execution as a directed acyclic graph $G=(V,E)$. Each node specifies a subgoal with a local evidence predicate; edges encode prerequisites. Nodes cover visual evidence, computation, memory operations, and the final answer. Let $b_{t,i}\in\{0,1\}$ indicate whether node $i$'s local predicate holds in the current state and recorded execution history, including its provenance conditions. Let $\mathrm{Anc}(i)$ contain all ancestors of $i$, excluding $i$ itself. Strict node verification is
\begin{equation}
 z_{t,i}=\prod_{j\in\{i\}\cup\mathrm{Anc}(i)} b_{t,j}.
 \label{eq:verified-node}
\end{equation}
For a root node, $z_{t,i}=b_{t,i}$. A non-root subgoal receives credit only when its evidence is valid and all prerequisites are satisfied. Later writes cannot alter an empty search result, so assigning credit requires a new search. A search that returned the required ID can become verified once its prerequisites hold.  
The trace records what happened; the edges determine when it counts. With positive node weights $w_i$, set to one by default, verified progress is
\begin{equation}
 \Phi_t=\frac{1}{W}\sum_{i\in V}w_i z_{t,i},\qquad W=\sum_{i\in V}w_i .
 \label{eq:strict-potential}
\end{equation}
Each task program derives its graph, evidence predicates and answer from source annotations and deterministic rules. An LLM assists with instruction wording. Replaying a reference trajectory checks executability. For an OCR write, the verifier checks the text, the source image and the successful tool call cited by the record.

\begin{table*}[t]
\centering
\caption{\textbf{Scale, task coverage, and evaluation capabilities.} Instances count benchmark candidates. Task families count top-level categories across the full suite, including all ten ViInteract execution domains. ViInteract combines visual interaction, intermediate progress scoring, and provenance- and prerequisite-checked evidence paths to evaluate complete task execution across multiple steps.}
\label{tab:benchmark-contract}
\small
\setlength{\tabcolsep}{4pt}\renewcommand{\arraystretch}{1.2}
\begin{tabularx}{\textwidth}{@{}l>{\raggedright\arraybackslash}X*{6}{>{\centering\arraybackslash}c}@{}}
\toprule
\multirow{2}{*}{\textbf{Dataset / suite}} & \multirow{2}{*}{\textbf{Tool interface}} & \multicolumn{2}{c}{\textbf{Scale and coverage}} & \multicolumn{4}{c}{\textbf{Evaluation capabilities}}\\
\cmidrule(lr){3-4}\cmidrule(lr){5-8}
 & & \shortstack{\textbf{Benchmark}\\\textbf{instances}} & \shortstack{\textbf{Suite task}\\\textbf{families}} & \shortstack{\textbf{Visual}\\\textbf{input}} & \shortstack{\textbf{Agent}\\\textbf{interaction}} & \shortstack{\textbf{Progress}\\\textbf{scoring}} & \shortstack{\textbf{Evidence-path}\\\textbf{verification}}\\
\midrule
Mantis-Eval~\cite{jiang2024mantis} & None & 217 & 4 & \checkmark & -- & -- & --\\
MuirBench~\cite{wang2024muirbench} & None & 2,600 & 12 & \checkmark & -- & -- & --\\
VisualWebArena~\cite{koh2024visualwebarena} & Browser tools & 910 & 3 & \checkmark & \checkmark & -- & --\\
OSWorld~\cite{xie2024osworld} & GUI tools & 369 & 5 & \checkmark & \checkmark & -- & --\\
AgentBoard~\cite{ma2024agentboard} & Environment tools & 1,013 & 9 & -- & \checkmark & \checkmark & --\\
\midrule
\rowcolor{resultshade}
\textbf{ViInteract (ours)} & \textbf{Visual tools} & \textbf{4,000} & \textbf{10} & \textbf{\checkmark} & \textbf{\checkmark} & \textbf{\checkmark} & \textbf{\checkmark}\\
\bottomrule
\end{tabularx}
\end{table*}

Strict completion requires every node, including the final answer, $d_t=\prod_i z_{t,i}=\prod_i b_{t,i}$. The program compiler enforces two structural rules, detailed in the supplement. The answer node is the unique sink: every evidence, computation, memory and retrieval node reaches it, so the answer predicate cites records whose producing events trace back through the graph. Every edge is enforced by provenance. For $j\to i$, the predicate $b_i$ cites an event produced by node $j$, so the order required for correctness is part of the predicate the product checks. An action therefore receives a node's credit when it supplies the missing cited evidence or completes the last missing prerequisite. The edges decide when that credit enters $\Phi_t$; Section~\ref{sec:reward-design} measures the effect.

\paragraph{Retrieval coverage and verified evidence.}
A search can recover a record before the verifier has established its provenance. We use the returned IDs to give the search immediate feedback while the verifier tracks whether the recovered evidence supports a dependent subgoal. For a retrieval node $i$, let $\mathcal M_i$ be its required record IDs and $\mathcal R_{t,i}$ the set returned through step $t$, after resolving aliases. Its coverage is $c_{t,i}=|\mathcal R_{t,i}\cap\mathcal M_i|/|\mathcal M_i|$. Set $c_{t,i}=0$ for other nodes or an empty required set. We use the same node weights as in Eq.~\ref{eq:strict-potential} to define the training potential
\begin{equation}
 \widetilde{\Phi}_t=\frac{1}{W}\sum_{i\in V} w_i\max\{z_{t,i},c_{t,i}\}.
 \label{eq:dense-potential}
\end{equation}
For binary $z$, $\max(z,c)=z+(1-z)c$: an unverified retrieval node earns credit for each required ID it recovers, and node $i$ never contributes more than $w_i/W$. Hence $\widetilde{\Phi}_t\geq\Phi_t$, with the gap confined to unverified retrieval nodes, and $\widetilde{\Phi}_t=\Phi_t$ on every non-retrieval node. Coverage enters training only; all reported metrics use $z$. Section~\ref{sec:reward-design} varies the weight on this auxiliary term.

\subsection{Net-Progress Reward}
\begin{table*}[!t]
\centering
\caption{\textbf{Main results on ViInteract.} API models are evaluated directly in (a) and with the shared Harness in (b). Qwen training stages use native agent execution in (c), where Answer and Strict use Core200 and the remaining four metrics use Process500. Scores are percentages ($\uparrow$). Bold marks the best value within each API group and each Qwen backbone.}
\label{tab:avm-main}\label{tab:api-harness}
\footnotesize
\setlength{\tabcolsep}{4pt}\renewcommand{\arraystretch}{1.12}
\begin{tabularx}{\textwidth}{
    @{}
    >{\raggedright\arraybackslash}p{0.30\textwidth}
    *{6}{>{\centering\arraybackslash}X}
    @{}}
\toprule
\multirow{2}{*}{\textbf{Model / Backbone}} & \multicolumn{6}{c}{\textbf{Evaluation metrics} $\uparrow$}\\
\cmidrule(lr){2-7}
 & \textbf{Answer} & \textbf{Strict} & \textbf{Graph} & \textbf{Tool} & \textbf{Protocol} & \textbf{State}\\
\midrule
\multicolumn{7}{@{}l}{\textit{(a) Direct API evaluation}}\\
\midrule
GPT-4o & 12.0 & 0.5 & 25.9 & 49.2 & 67.4 & 67.4\\
Claude Opus 4.6 & 18.0 & \textbf{11.5} & \textbf{37.6} & 54.6 & 95.9 & 95.6\\
Qwen3-VL-Plus & 3.0 & 2.5 & 16.0 & 19.3 & 49.3 & 49.3\\
Qwen3.8-Flash & 9.5 & 8.5 & 33.2 & 41.6 & 54.1 & 54.0\\
Gemini 3.8 Flash & \textbf{20.5} & 2.5 & 29.0 & 39.8 & 92.2 & 92.2\\
Kimi K2.5 & 13.5 & 7.0 & 37.4 & \textbf{55.3} & \textbf{96.3} & \textbf{96.3}\\
\midrule
\multicolumn{7}{@{}l}{\textit{(b) API models with Harness}}\\
\midrule
GPT-4o & 35.0 & \textbf{30.5} & \textbf{41.3} & \textbf{69.6} & 95.3 & 99.9\\
Claude Opus 4.6 & 30.5 & 26.5 & 35.2 & 57.8 & 86.1 & \textbf{100.0}\\
Qwen3-VL-Plus & 31.5 & 18.5 & 29.7 & 64.1 & 85.9 & 99.9\\
Qwen3.8-Flash & \textbf{35.5} & 19.5 & 33.5 & 66.6 & 98.0 & 99.7\\
Gemini 3.8 Flash & 34.5 & 2.5 & 21.5 & 58.3 & \textbf{99.3} & 99.9\\
Kimi K2.5 & 22.0 & 14.5 & 31.7 & 42.2 & 66.3 & \textbf{100.0}\\
\midrule
\multicolumn{7}{@{}l}{\textit{(c) Qwen2.5-VL training-stage comparison}}\\
\midrule
Qwen2.5-VL-3B Base & 0.0 & 0.0 & 11.5 & 40.6 & 68.9 & \textbf{88.8}\\
Qwen2.5-VL-3B SFT & 19.0 & 15.0 & 29.5 & 55.5 & 93.1 & 73.3\\
Qwen2.5-VL-3B Standard GRPO & 20.0 & 15.5 & 32.7 & 58.3 & 92.4 & 75.0\\
\rowcolor{resultshade}
\textbf{Qwen2.5-VL-3B ViStep-RL} & \textbf{24.0} & \textbf{20.5} & \textbf{51.8} & \textbf{75.6} & \textbf{97.6} & 82.4\\
\midrule
Qwen2.5-VL-7B Base & 12.0 & 10.0 & 9.3 & 16.4 & 61.5 & \textbf{94.8}\\
Qwen2.5-VL-7B SFT & 55.0 & 46.5 & 42.0 & 57.9 & 62.4 & 61.3\\
Qwen2.5-VL-7B Standard GRPO & 58.0 & 48.5 & 45.8 & 64.2 & 69.0 & 64.7\\
\rowcolor{resultshade}
\textbf{Qwen2.5-VL-7B ViStep-RL} & \textbf{67.5} & \textbf{60.0} & \textbf{61.6} & \textbf{86.7} & \textbf{91.2} & 91.0\\
\bottomrule
\end{tabularx}
\end{table*}

An agent may retain the same observation across several actions. We reward the transition that changes its contribution to the task~\cite{ng1999shaping}. Let $e_t$ indicate a protocol violation, failed state or tool check, or an incorrect or unsupported answer on action $t$. The charge for an erroneous action equals the mean node share, $1/|V|$, so a mistake costs as much as one average subgoal on that graph. The transition reward is
\begin{equation}
 r_t=\underbrace{\widetilde{\Phi}_t-\widetilde{\Phi}_{t-1}}_{\delta_t\,\text{ (progress)}}-\frac{e_t}{|V|}.
 \label{eq:trajectory-return}
\end{equation}
A valid answer earns credit through its own node. Summing over an episode gives
\begin{equation}
 R(\tau)=\sum_{t=1}^T r_t
 =\widetilde{\Phi}_T-\widetilde{\Phi}_0-\frac{1}{|V|}\sum_{t=1}^T e_t.
 \label{eq:telescoping}
\end{equation}
Holding a record unchanged earns no further progress. A cumulative reward $\sum_t\widetilde{\Phi}_t$ pays for the same record at every later step: a node verified at step $t_i$ and kept verified contributes $w_i(T-t_i+1)/W$, whereas Eq.~\ref{eq:telescoping} pays $w_i/W$ once. With the action budget fixed, reaching a subgoal earlier leaves more actions for the nodes that remain.

\begin{proposition}[Shaping consistency]
\label{prop:shaping}
(i) $\Phi_t=1$ if and only if $d_t=1$, and, because the answer node is the unique sink, $\widetilde{\Phi}_t=1$ if and only if $d_t=1$. (ii) Losing a node's verification and later regaining it earns zero net progress over that cycle, while every error inside it is still charged. (iii) $0\leq\widetilde{\Phi}_t-\Phi_t\leq W_{\mathrm{ret}}/W$, where $W_{\mathrm{ret}}$ sums the retrieval-node weights, and any trace with $d_T=0$ ends at most $1-w_{\mathrm{ans}}/W$ on the progress term.
\end{proposition}
The supplement proves these claims. Full training credit coincides with Strict completion, a re-verification cycle adds no progress, and auxiliary retrieval credit remains strictly below a fully verified trace on the progress term.

\subsection{Graph-Guided Policy Optimization}
For a given prompt, we retain groups whose rollout returns vary.
We compare actions from the same prompt, execution phase and visual turn, binning pre-action progress $\widetilde{\Phi}_{t-1}$ to the nearest tenth. Actions in a bin can have different evidence histories; the bin supplies a stage-specific reward scale. Let $u_t$ be the centered, standardized reward in this bin, with $u_t=0$ at zero variance. Centering alone can give a failed action a positive score when its peers fare worse. We use the transition reward to set the update direction and the within-bin comparison to adjust its magnitude. Writing $s_t=\operatorname{sign}(r_t)$ with $\operatorname{sign}(0)=0$, the signed action weight is
\begin{equation}
 A_t=s_t[s_t u_t]_+ + g(r_t),
 \label{eq:group-advantage}
\end{equation}
where $[x]_+=\max(x,0)$ and $g(r)=\tfrac14\operatorname{clip}(20r,-1,1)$. The clipped term gives feedback even in a singleton bin. It saturates at $\pm\tfrac14$, below the unit scale of $u_t$. The relative term amplifies positive rewards above the bin mean and negative rewards below it. Since $r\,g(r)\geq0$, multiplying Eq.~\ref{eq:group-advantage} by $r_t$ gives $|r_t|[s_tu_t]_++r_tg(r_t)\geq0$. A failed action therefore receives no positive weight, even when every peer does worse.

Each update keeps the actions with the largest $|A_t|$, up to two per rollout at 3B and four at 7B. All tokens of a selected action share that action's weight in a clipped policy loss regularized toward the SFT reference. Section~\ref{sec:main-results} compares this action-level update with a trajectory-level update. The supplement gives thresholds and the token loss.


\begin{table*}[t]
    \centering
    \small
    \setlength{\tabcolsep}{3pt}
    \renewcommand{\arraystretch}{1.18}
    \caption{\textbf{Ablation of progress rewards and verification constraints.}
    Progress specifies the training reward. The two gate columns indicate where prerequisites apply during training.
    All variants share matched training budgets and are
    scored with the same full verifier on identical episodes within each panel.
    Strict and Graph are percentages; higher is better.}
    \label{tab:reward-ablations}
    
    \begin{tabularx}{\textwidth}{
        @{}
        >{\raggedright\arraybackslash}p{0.25\textwidth}
        ccc
        *{4}{>{\centering\arraybackslash}X}
        @{}}
    \toprule
    \multirow{2}{*}{\textbf{Training variant}}
    & \multicolumn{3}{c}{\textbf{Design components}}
    & \multicolumn{2}{c}{\textbf{Process500}}
    & \multicolumn{2}{c}{\textbf{Holdout200}} \\
    \cmidrule(lr){2-4}
    \cmidrule(lr){5-6}
    \cmidrule(lr){7-8}
    & Progress
    & Other nodes
    & Answer node
    & Strict $\uparrow$
    & Graph $\uparrow$
    & Strict $\uparrow$
    & Graph $\uparrow$ \\
    \midrule
    
    Cumulative reward
    & Cumulative
    & $\checkmark$
    & $\checkmark$
    & 22.6 & 49.4 & 12.0 & 33.6 \\
    
    No progress reward
    & None
    & $\checkmark$
    & $\checkmark$
    & 23.8 & \textbf{54.1} & 12.5 & 33.6 \\
    
    No prerequisite gating
    & Incremental
    & $\times$
    & $\times$
    & 25.2 & 51.7 & 14.5 & 35.0 \\
    
    No answer prerequisite gate
    & Incremental
    & $\checkmark$
    & $\times$
    & 23.4 & 51.4 & 14.5 & 35.3 \\
    
    \midrule
    \rowcolor{resultshade}
    \textbf{ViStep-RL}
    & \textbf{Incremental}
    & $\checkmark$
    & $\checkmark$
    & \textbf{25.8}
    & 51.8
    & \textbf{15.5}
    & \textbf{36.3} \\
    \bottomrule
    \end{tabularx}
    \end{table*}
    
\section{Experiments}
\label{sec:experiments}

\subsection{Benchmark and Evaluation Protocol}
\textbf{ViInteract} spans ten execution domains and 34 task-program types. It comprises 76,000 SFT episodes, a 15,000-episode RL source pool, and 4,000 benchmark candidates. The benchmark pool covers nine of the ten domains and 29 program types. Its primary evaluation set, Core200, contains 200 frozen episodes with an application-weighted composition. Process500 (500 episodes) measures execution behavior, and Holdout200 (200 episodes, selected without model predictions) tests fresh instances. Table~\ref{tab:benchmark-contract} compares the evaluation contract with established visual and interactive suites; the supplement gives the taxonomy and panel composition.

Each task pairs visual inputs with reference evidence, prerequisite relations and an answer; the agent sees only the current image and public observations. Extending AgentBoard~\cite{ma2024agentboard}, ViInteract tests whether an observation stays usable after its image leaves view.

\paragraph{Metrics.}
\emph{Answer} scores the final response; \emph{Strict} requires every evidence predicate. \emph{Graph} measures the required-node verification rate, \emph{Tool} the tool-success rate, \emph{Protocol} the schema-valid action rate, and \emph{State} the accepted transition rate. In Table~\ref{tab:api-harness}, direct API runs use a separate matched panel and Harness runs use Core200. For native Qwen, Answer and Strict use Core200; Graph, Tool, Protocol and State use Process500. Paired intervals resample shared episode IDs.

\paragraph{Training and evaluation controls.}
We evaluate Qwen2.5-VL~\cite{Bai2025Qwen25VLTR} at 3B and 7B.
We adapt attention and MLP projections with LoRA and evaluate each policy under fixed decoding settings. Qwen2.5-VL-7B training consists of 1,000 updates over 32,000 rollouts and takes 36.94 GPU-hours on four GPUs. The release audit found no shared episode IDs, complete image paths, or source-sequence IDs between training and the three evaluation panels. The supplement gives implementation details.

\subsection{Post-Training and General-Purpose Models}
\label{sec:main-results}
\begin{figure*}[t]
\centering
\includegraphics[width=\textwidth]{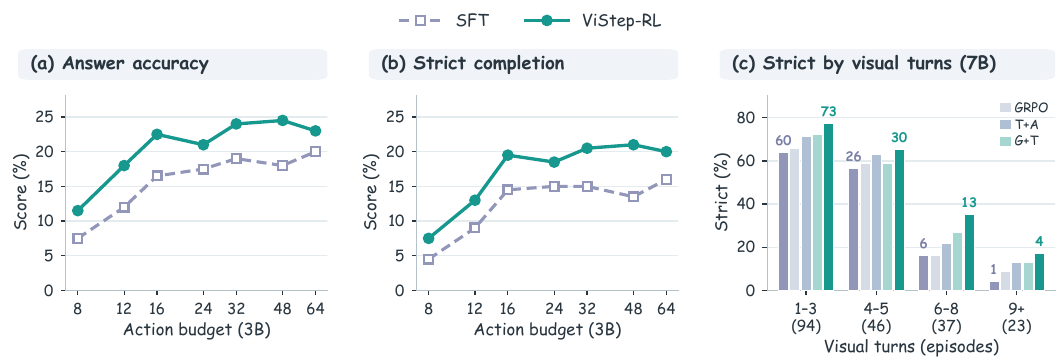}
\caption{\textbf{Action budget and visual horizon.} (a, b) Qwen2.5-VL-3B SFT and ViStep-RL on 200 episodes per budget. (c) Qwen2.5-VL-7B Strict completion on Core200, grouped by visual turns; episode counts are shown below group labels, and numbers above the SFT and ViStep-RL bars give completed episodes. T+A uses terminal reward with action updates; G+T uses graph reward with trajectory updates.}\label{fig:action-budget}
\end{figure*}
\paragraph{Task-graph training improves completion at both Qwen scales.}
On Core200, Qwen2.5-VL-7B completes 120 of 200 episodes under the Strict criterion after ViStep-RL, up from 93 after SFT. The gain is 13.5 percentage points (paired 95\% bootstrap interval [8.0, 19.0]). The 3B model also improves, from 15.0\% to 20.5\% Strict (Table~\ref{tab:api-harness}(c)).

Standard GRPO provides a matched reinforcement-learning control. Both methods start from the same SFT policy and use the same tools, action budget, rollouts and optimizer updates. Standard GRPO uses final-answer correctness as its reward and shares one group-relative advantage across each trajectory; it updates every action, ViStep-RL at most four per 7B rollout. ViStep-RL improves Strict from 15.5\% to 20.5\% at 3B and from 48.5\% to 60.0\% at 7B. The corresponding Answer gains are 4.0 and 9.5 points. On Process500, Graph improves by 19.1 and 15.8 points, and Tool by 17.3 and 22.5 points. Crossing the two design choices at 7B helps isolate their respective contributions. With action-level updates fixed, graph reward raises Strict from 53.5\% to 60.0\%. With graph reward fixed, action-level updates raise it from 54.0\% to 60.0\%. Section~\ref{sec:reward-design} separates the components of the graph reward.

The 7B Core200 traces show a change in \emph{which} actions succeed: from SFT to ViStep-RL, required writes completed rise from 46.24\% to 54.67\% and retrievals from 60.51\% to 69.99\%, while mean actions fall from 15.76 to 15.00.

\paragraph{General-purpose models under the shared Harness.}
With shared public prompts, tools and scorer on Core200, six API models reach 22.0--35.5\% Answer; Qwen2.5-VL-7B scores 16.5\% at Base and 59.5\% after task training. The API models achieve 99.7--100\% state-transition acceptance, yet only 2.5--30.5\% of their episodes satisfy Strict. Strict thus separates valid execution from evidence-supported answers.

\subsection{Reward Design: Incremental Task Progress}
\label{sec:reward-design}
All policies in Table~\ref{tab:reward-ablations} share the same Qwen2.5-VL-3B initialization, training budget and action-level update. The first row tests Eq.~\ref{eq:telescoping}: the cumulative variant instead rewards the current potential at every step. Rewarding its change gives 25.8\% Strict on Process500 versus 22.6\% for cumulative reward (paired 95\% interval for the gain [1.0, 5.6] points, McNemar $p=0.0113$). On Holdout200, the same comparison similarly gives 15.5\% versus 12.0\%.

The remaining rows isolate the progress measure. Removing the progress term leaves the execution cost in Eq.~\ref{eq:trajectory-return} and drops Strict from 25.8\% to 23.8\% on Process500. Every change to the graph lowers Strict on Process500: removing all prerequisite gates gives 25.2\%, a separately trained random-DAG control 24.6\%, and removing only the answer node's gate 23.4\%. Both gate removals reach 14.5\% Strict on Holdout200 against 15.5\% for ViStep-RL, and the full graph gives the highest Holdout200 Graph score, 36.3\%. The largest single effect is the switch from cumulative to incremental reward, which Proposition~\ref{prop:shaping}(ii) predicts: the cumulative form pays again for every node it re-verifies.

The coverage sweep tests the effect of the auxiliary term in Eq.~\ref{eq:dense-potential}. Giving unverified retrieval nodes zero coverage credit yields 122/500 Strict successes on Process500 and 26/200 on Holdout200. With full coverage credit, the counts rise to 129 and 31. Proposition~\ref{prop:shaping}(iii) bounds this auxiliary credit below a fully verified trace.

\begin{figure*}[t]
\centering
\includegraphics[width=\textwidth]{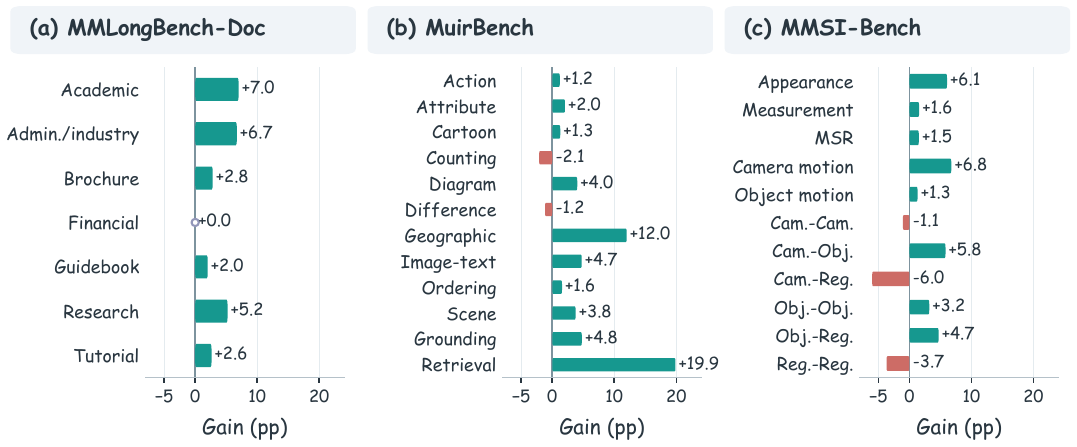}
\caption{\textbf{Category-wise performance on external benchmarks.}
Accuracy changes for Qwen2.5-VL-3B ViStep-RL relative to matched references on (a) MMLongBench-Doc, (b) MuirBench and (c) MMSI-Bench. Teal bars indicate improvements, while coral bars indicate declines. Cam., Obj., and Reg. denote camera, object, and region.}
\label{fig:external-categories}
\end{figure*}

\subsection{Memory: Evidence Selection and Persistence}
\begin{table}[t]
    \centering
    \small
    \setlength{\tabcolsep}{3pt}
    \renewcommand{\arraystretch}{1.12}
    \caption{\textbf{Effects of memory and evidence interventions.}
(a) Fixed 3B policy and memory budget on Core200. (b) Holdout200; Answer scores final responses independently of evidence checks.}
    \label{tab:memory-analysis}
    \begin{tabular*}{\columnwidth}{@{\extracolsep{\fill}}lrrrr@{}}
    \toprule
    \multicolumn{5}{l}{\textit{(a) Writing $\times$ retrieval, $N=200$/cell}} \\
    Write / retrieval & Answer & Strict & Graph & Tool \\
    \midrule
    Active / Active
    & \textbf{24.5} & \textbf{17.0} & \textbf{49.0} & 69.6 \\
    Active / BM25
    & 22.0 & 14.5 & 48.5 & \textbf{70.8} \\
    Automatic / Active
    & 16.5 & 8.5 & 43.2 & 60.4 \\
    Automatic / BM25
    & 15.0 & 7.5 & 43.6 & 62.0 \\
    \midrule
    \multicolumn{5}{l}{\textit{(b) Evidence interventions, Holdout200}} \\
    Intervention & Answer & Strict & Graph & Tool \\
    \midrule
    ViStep-RL
    & \textbf{21.0} & \textbf{15.5} & \textbf{36.33} & \textbf{66.33} \\
    Disable memory writes
    & 0.0 & 0.0 & 21.45 & 34.32 \\
    Corrupt stored memory
    & 3.5 & 0.0 & 21.65 & 60.47 \\
    Remove required tools
    & 6.5 & 4.5 & 17.84 & 33.09 \\
    \bottomrule
    \end{tabular*}
    \end{table}
    
\begin{table}[!t]
\centering\small
\caption{\textbf{Answer quality and cost on Core200.} Both 3B policies receive the same public inputs. Direct supplies observation notes; Harness also provides a tool controller. Native execution is evaluated in Table~\ref{tab:api-harness}(c). Costs are per episode.}
\label{tab:budget-framework}
\setlength{\tabcolsep}{3pt}
\begin{tabular*}{\columnwidth}{@{\extracolsep{\fill}}lrrrr@{}}
\toprule
 & \multicolumn{2}{c}{Base} & \multicolumn{2}{c}{ViStep-RL}\\
\cmidrule(lr){2-3}\cmidrule(l){4-5}
 & Direct & Harness & Direct & Harness\\
\midrule
Answer (\%) & 16.5 & 18.0 & 18.0 & \textbf{29.0}\\
Parse fail (\%) & 4.5 & 3.0 & 5.0 & 0.5\\
\midrule
Input tokens & 2{,}798.17 & 22{,}197.81 & 2{,}796.97 & 24{,}249.02\\
Output tokens & 324.58 & 703.60 & 455.61 & 903.83\\
Calls & 4.815 & 15.850 & 4.815 & 15.235\\
Time (s) & 13.38 & 30.38 & 23.58 & 37.00\\
\bottomrule\end{tabular*}
\end{table}

For the memory analysis, Table~\ref{tab:memory-analysis}(a) fixes the same 3B weights and Core200 IDs as the native comparison, with all four conditions sharing a bounded store: eight records, 512 serialized characters per record, and four returns per retrieval. Choosing what to write improves Answer by 8.0 points under learned retrieval and by 7.0 under BM25. Averaged over retrievers, active writing gains 7.75 points in Strict; averaged over writing modes, learned retrieval gains 1.75 points over BM25.

On Holdout200, the intact policy answers 42 correctly, compared with zero when writes are disabled and seven when stored records are corrupted (Table~\ref{tab:memory-analysis}(b)). Strict rejects those seven because their evidence chains fail. Under corruption, 60.47\% of tool calls still succeed: a successful call can return a record that no longer supports the answer.

\begin{figure}[t]
\centering
\includegraphics[width=\linewidth]{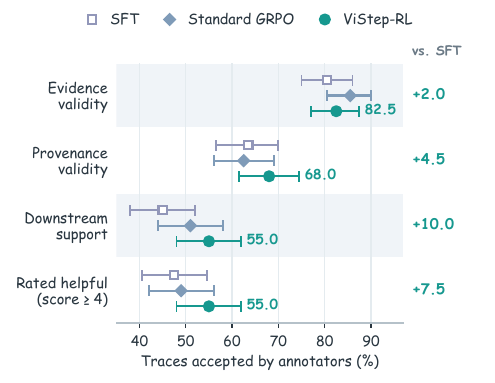}
\caption{\textbf{Human evaluation of stored evidence} on 200 paired Core200 episodes (7B, three annotators, majority vote). Error bars are 95\% bootstrap intervals; the right column is the gain over SFT.}
\label{fig:human-evidence}
\end{figure}

\paragraph{Human evaluation.}
Three annotators rated the 7B traces of all 200 Core200 episodes for SFT, Standard GRPO and ViStep-RL, with model names hidden and order randomized. Before seeing the final answer, each judged whether a record matched its source image, whether it named that image and the producing tool event, and whether the retrieved records sufficed for a later decision. A trace counts when at least two annotators accept it. In Figure~\ref{fig:human-evidence}, ViStep-RL reaches 68.0\% provenance validity and 55.0\% downstream support, against 63.5\% and 45.0\% for SFT and 62.5\% and 51.0\% for Standard GRPO. Evidence validity stays within 80.5--85.5\% across policies, which share one visual backbone. Mean helpfulness rises from 3.09 and 3.10 to 3.24. The supplement gives rating instructions and agreement.

\subsection{Action Budget and Model--Harness Interaction}
Figure~\ref{fig:action-budget}(a, b) varies the action cap. Across all seven budgets, ViStep-RL exceeds SFT by 3.0--6.5 points in Answer and 3.0--7.5 in Strict. Strict peaks at 21.0\% at budget 48.

\paragraph{Visual horizon.}
Grouping 7B Core200 episodes by images shown, Figure~\ref{fig:action-budget}(c) shows completions rising from 6 to 13 at 6--8 turns and from 1 to 4 at nine or more turns.

Table~\ref{tab:budget-framework} compares Direct observation notes with tool-enabled Harness execution on identical Core200 IDs. Harness raises Answer by 1.5 points at Base and 11.0 after training. ViStep-RL uses $8.67\times$ the input tokens and $3.16\times$ the calls, with latency rising from 23.58 to 37.00 seconds per episode.

\subsection{Generalization to New Instances and Tasks}

\paragraph{New visual instances.}
Holdout200 excludes Core200 and earlier panels and is selected without model predictions. SFT-to-ViStep-RL Strict rises from 8.0\% to 15.5\% at 3B and from 40.0\% to 50.5\% at 7B.

\paragraph{External visual reasoning.}
Figure~\ref{fig:external-categories} compares Qwen2.5-VL-3B ViStep-RL with its reference under matched inputs and prompts. MMLongBench-Doc~\cite{ma2024mmlongbenchdoc} Accuracy rises from 15.00\% to 19.10\%, with a 27.05-point gain on unanswerable questions. MuirBench~\cite{wang2024muirbench} rises from 35.62\% to 40.15\%, gaining 19.86 points in retrieval and 12.00 in geographic reasoning. MMSI-Bench~\cite{yang2026mmsibench} rises from 25.50\% to 27.20\%.

\section{Conclusion}
\label{sec:conclusion}
ViStep-RL trains visual agents to retain and recover evidence across visual turns through graph-based progress rewards and action-level updates. Under matched interaction resources, 7B Strict rises from 48.5\% to 60.0\% over terminal-answer GRPO. Selective writing improves Answer under a fixed memory budget, while human evaluation shows gains in provenance and downstream support.

\clearpage
\section*{LLM Usage Statement}
OpenAI Codex assisted with manuscript drafting, language editing, literature checks, and preparation of analysis scripts. LLMs also assisted with refining the wording of dataset instructions; task graphs and answer labels were determined by programmatic rules.\par

{\small\bibliographystyle{ieeenat_fullname}\bibliography{references}}
\clearpage
\appendix
\setlength{\textfloatsep}{12pt plus 2pt minus 2pt}
\setlength{\dbltextfloatsep}{12pt plus 2pt minus 2pt}
\setlength{\floatsep}{10pt plus 2pt minus 2pt}
\setlength{\dblfloatsep}{10pt plus 2pt minus 2pt}
\raggedbottom
\setcounter{table}{0}
\setcounter{figure}{0}
\setcounter{equation}{0}
\setcounter{algorithm}{0}
\setcounter{suppthm}{0}
\renewcommand{\theHtable}{supp.\arabic{table}}
\renewcommand{\theHfigure}{supp.\arabic{figure}}
\renewcommand{\theHequation}{supp.\arabic{equation}}
\twocolumn[{\centering
{\Large\bfseries ViStep-RL: Stepwise Reinforcement Learning for\\
Long-Horizon Visual Agents\par}
\vspace{5pt}
{\large\bfseries Supplementary Material\par}
\vspace{14pt}}]

\renewcommand{\thetable}{S\arabic{table}}
\renewcommand{\thefigure}{S\arabic{figure}}
\renewcommand{\theequation}{S\arabic{equation}}
This supplement develops the theory behind ViStep-RL, including the proof of Proposition~\ref{prop:shaping} and the training procedure. It then gives the benchmark card and training recipe, the Core200 execution analysis, the human evaluation protocol, a qualitative trajectory comparison and the general-purpose model evaluation protocol. All scores are percentages unless stated otherwise.

\section{Theoretical Analysis}
\label{sec:supp-theory}
This section states the structural assumptions on a task program, characterizes the verified set, and proves that the net-progress return and the signed action weight are consistent with Strict completion. Algorithm~\ref{alg:vistep} summarizes one training iteration.

\paragraph{Notation and assumptions.}
Write $B_t=\{i:b_{t,i}=1\}$ and $Z_t=\{i:z_{t,i}=1\}$ for the locally satisfied and the verified nodes. A set $S\subseteq V$ is \emph{ancestor-closed} if $i\in S$ implies $\mathrm{Anc}(i)\subseteq S$. Let $D_j=\{j\}\cup\mathrm{Desc}(j)$, and let $\omega(S)=\sum_{i\in S}w_i/W$. The ViInteract program compiler (Section~\ref{sec:supp-graph-construction}) builds every task program with the following properties:
\begin{enumerate}[label=(A\arabic*),leftmargin=*,itemsep=1pt,topsep=2pt]
\item node weights are positive;
\item no node is verified and no record is returned at $t=0$, so $\widetilde{\Phi}_0=\Phi_0=0$;
\item the answer node $a$ is the unique sink of $G$, and $c_{t,a}=0$;
\item every edge $j\to i$ is a provenance condition of $b_i$: the evidence of node $i$ cites an event produced by node $j$.
\end{enumerate}
Under (A3), $\mathrm{Anc}(a)=V\setminus\{a\}$. Indeed, following outgoing edges from any node $i\neq a$ must terminate because $G$ is finite and acyclic, and it can only terminate at a node without outgoing edges, which is $a$. Let $\alpha=w_a/W$ denote the answer share.

\begin{algorithm}[t]
\caption{One ViStep-RL training iteration}
\label{alg:vistep}
\small
\begin{algorithmic}[1]
\Require policy $\pi_\theta$, SFT reference $\pi_{\mathrm{ref}}$, prompts $\mathcal P$, group size $n$, actions kept per rollout $k$
\For{each prompt $q\in\mathcal P$}
 \State sample rollouts $\tau^{(1)},\ldots,\tau^{(n)}\sim\pi_\theta$ in the runtime
 \For{each transition $t$ of each rollout}
  \State $Z_t\gets\textsc{Verify}(B_t)$ \Comment{Lemma~\ref{lem:closure}}
  \State compute $\widetilde{\Phi}_t$ by Eq.~\ref{eq:dense-potential} and $r_t$ by Eq.~\ref{eq:trajectory-return}
 \EndFor
 \If{$\operatorname{std}_g R(\tau^{(g)})\leq10^{-5}$}
  \State skip $q$
 \EndIf
 \State bin actions by phase, visual turn, and $\operatorname{round}(10\widetilde{\Phi}_{t-1})$
 \State $u_t\gets$ standardized $r_t$ within its bin
 \State $A_t\gets s_t[s_tu_t]_+ + g(r_t)$ \Comment{Eq.~\ref{eq:group-advantage}}
 \State keep the $k$ actions of each rollout with largest $|A_t|$
\EndFor
\State update $\theta$ with the clipped token loss of Eq.~\ref{eq:supp-policy-objective}
\Statex
\Function{Verify}{$B$}
 \State $Z\gets\emptyset$
 \Repeat
  \State add every $i\in B\setminus Z$ whose parents all lie in $Z$
 \Until{$Z$ is unchanged}
 \State \Return $Z$
\EndFunction
\end{algorithmic}
\end{algorithm}

\subsection{The Verified Set}
\begin{supplemma}[Closure]
\label{lem:closure}
$Z_t$ is the largest ancestor-closed subset of $B_t$, and \textsc{Verify} returns it.
\end{supplemma}
\begin{proof}
By Eq.~\ref{eq:verified-node}, $i\in Z_t$ if and only if $\{i\}\cup\mathrm{Anc}(i)\subseteq B_t$. Hence $Z_t\subseteq B_t$. If $i\in Z_t$ and $j\in\mathrm{Anc}(i)$, then $\{j\}\cup\mathrm{Anc}(j)\subseteq\{i\}\cup\mathrm{Anc}(i)\subseteq B_t$, so $j\in Z_t$ and $Z_t$ is ancestor-closed. If $S\subseteq B_t$ is ancestor-closed and $i\in S$, then $\{i\}\cup\mathrm{Anc}(i)\subseteq S\subseteq B_t$, so $S\subseteq Z_t$.

Let $Y$ be the output of \textsc{Verify}. Every node added to $Y$ lies in $B_t$ and has its parents in $Y$; by induction on the order of addition, $Y\subseteq Z_t$. Suppose $Z_t\setminus Y\neq\emptyset$ and choose $i$ in it that is earliest in a topological order. Every parent of $i$ lies in $Z_t$ and precedes $i$, so it lies in $Y$. Since $b_{t,i}=1$, \textsc{Verify} would add $i$ before terminating, a contradiction. Hence $Y=Z_t$.
\end{proof}

\begin{supplemma}[Counterfactual credit]
\label{lem:credit}
Suppose action $t$ changes only the predicate of node $j$.
If $b_j$ changes from $1$ to $0$, then $Z_t=Z_{t-1}\setminus D_j$ and $\Phi_{t-1}-\Phi_t=\omega(Z_{t-1}\cap D_j)$.
If $b_j$ changes from $0$ to $1$, then $Z_t=Z_{t-1}\cup U_j$ with $U_j=\{i\in D_j:\{i\}\cup\mathrm{Anc}(i)\subseteq B_t\}$, and $\Phi_t-\Phi_{t-1}=\omega(U_j)$.
\end{supplemma}
\begin{proof}
\emph{Loss.} Here $B_t=B_{t-1}\setminus\{j\}$. The set $S=Z_{t-1}\setminus D_j$ lies in $B_t$. It is ancestor-closed: if $i\in S$ had an ancestor $k\in D_j$, then $j\in\{k\}\cup\mathrm{Anc}(k)\subseteq\mathrm{Anc}(i)$, so $i\in D_j$, a contradiction. Conversely, any ancestor-closed $S'\subseteq B_t$ lies in $Z_{t-1}$ by Lemma~\ref{lem:closure}, and it contains no $i\in D_j$, since otherwise $j\in\{i\}\cup\mathrm{Anc}(i)\subseteq S'\subseteq B_t$. By Lemma~\ref{lem:closure}, $Z_t=S$.

\emph{Gain.} For $i\notin D_j$, the set $\{i\}\cup\mathrm{Anc}(i)$ does not contain $j$, so the status of $i$ is unchanged. Every $i\in D_j$ was unverified at $t-1$ because $j\notin B_{t-1}$, and it is verified at $t$ exactly when $i\in U_j$.
\end{proof}
Lemma~\ref{lem:credit} identifies the progress of an action with the weight of the subgoals whose verification would differ had the action not occurred. Evidence that a descendant gathered earlier is credited at the action that completes its last missing prerequisite, since that action is the one that makes the descendant verifiable.

\begin{suppcor}[Gating prices downstream evidence]
\label{cor:gating}
Let $\Phi^{b}_t=\omega(B_t)$ be the ungated potential. When an action invalidates a verified node $j$,
\[
 \Phi_{t-1}-\Phi_t=\omega(Z_{t-1}\cap D_j)\;\geq\;\frac{w_j}{W}=\Phi^{b}_{t-1}-\Phi^{b}_t,
\]
with equality if and only if $j$ has no verified descendant.
\end{suppcor}
\begin{proof}
The equality on the left is Lemma~\ref{lem:credit}; the one on the right holds because only $b_j$ changes. Since $j\in Z_{t-1}\cap D_j$ and all weights are positive, the inequality holds, and it is strict exactly when $Z_{t-1}\cap\mathrm{Desc}(j)\neq\emptyset$.
\end{proof}
Under the ungated potential, corrupting a record that supports $m$ verified descendants costs one node share and leaves the descendants credited although their source no longer holds. The gated potential charges the full supported subtree. This is the property that the memory-corruption and write-disable interventions in the main paper exercise.

\subsection{Proof of Proposition~\ref{prop:shaping}}
\label{sec:supp-shaping-proof}
\emph{Part (i).} If $d_t=1$, every $b_{t,i}=1$, so $Z_t=V$ and $\Phi_t=1$. Conversely, $\Phi_t=1$ with positive weights forces $Z_t=V$, hence $d_t=1$. For the training potential, $\widetilde{\Phi}_t=1$ forces $\max\{z_{t,i},c_{t,i}\}=1$ for every node. By (A3), $c_{t,a}=0$, so $z_{t,a}=1$. Because $\mathrm{Anc}(a)=V\setminus\{a\}$, $z_{t,a}=\prod_{j\in V}b_{t,j}=d_t$, so $d_t=1$. The converse follows from $\widetilde{\Phi}_t\geq\Phi_t$.

\emph{Part (ii).} Summing Eq.~\ref{eq:trajectory-return} over $[t_1,t_2]$ gives $\sum_{t=t_1}^{t_2}r_t=\widetilde{\Phi}_{t_2}-\widetilde{\Phi}_{t_1-1}-\frac{1}{|V|}\sum_{t=t_1}^{t_2}e_t$. If the state returns to its earlier verified set and coverage, the progress term vanishes, and each error inside the cycle still contributes $1/|V|$.

\emph{Part (iii).} For binary $z$, $\max\{z,c\}-z=(1-z)c$, which is zero on non-retrieval nodes and lies in $[0,1]$ on retrieval nodes. Summing with the node weights gives $0\leq\widetilde{\Phi}_t-\Phi_t\leq W_{\mathrm{ret}}/W$. If $d_T=0$, then $z_{T,a}=d_T=0$ and $c_{T,a}=0$, so the answer node contributes nothing and $\widetilde{\Phi}_T\leq1-\alpha$.\hfill$\square$

\subsection{Return Maximizers are Strict Completions}
\begin{suppthm}
\label{thm:maximizer}
Every trajectory satisfies $R(\tau)\leq1$, with equality if and only if $d_T=1$ and $\sum_te_t=0$. Every trajectory with $d_T=0$ satisfies $R(\tau)\leq1-\alpha$.
\end{suppthm}
\begin{proof}
By Eq.~\ref{eq:telescoping} and (A2), $R(\tau)=\widetilde{\Phi}_T-\frac{1}{|V|}\sum_te_t\leq1$. If $d_T=1$ and no error occurs, Proposition~\ref{prop:shaping}(i) gives $\widetilde{\Phi}_T=1$ and $R(\tau)=1$. Conversely, suppose $R(\tau)=1$ while $d_T=0$ or some $e_t=1$. Since $\widetilde{\Phi}_T\leq1$ and the error term is nonnegative, $R(\tau)=1$ forces both $\widetilde{\Phi}_T=1$ and $\sum_te_t=0$. Proposition~\ref{prop:shaping}(i) then gives $d_T=1$, a contradiction. The second claim is Proposition~\ref{prop:shaping}(iii) together with $e_t\geq0$.
\end{proof}

\begin{suppcor}
\label{cor:strict-bound}
For every policy $\pi$, $\Pr_\pi(d_T=1)\geq1-\bigl(1-\mathbb E_\pi[R]\bigr)/\alpha$.
\end{suppcor}
\begin{proof}
Let $p=\Pr_\pi(d_T=1)$. By Theorem~\ref{thm:maximizer}, $\mathbb E_\pi[R]\leq p+(1-p)(1-\alpha)=1-\alpha(1-p)$. Rearranging gives the bound.
\end{proof}
The expected return therefore controls the Strict completion rate. A policy cannot approach the maximal return while leaving a fixed fraction of episodes incomplete.

\subsection{Potential Differences are the Loop-Neutral Rewards}
\begin{suppdef}
A progress reward $F:[0,1]^2\to\mathbb R$ pays $F(\widetilde{\Phi}_{t-1},\widetilde{\Phi}_t)$ at transition $t$. It is \emph{loop-neutral} if $\sum_{k=1}^{K}F(\varphi_{k-1},\varphi_k)=0$ for every closed sequence $\varphi_0,\ldots,\varphi_K=\varphi_0$ in $[0,1]$.
\end{suppdef}
Loop neutrality states that holding, losing, and regaining the same evidence earns nothing. A reward that violates it pays a policy for repeating some cycle of verification states.

\begin{suppthm}
\label{thm:loop}
$F$ is loop-neutral if and only if $F(\varphi,\psi)=h(\psi)-h(\varphi)$ for some $h:[0,1]\to\mathbb R$.
\end{suppthm}
\begin{proof}
If $F(\varphi,\psi)=h(\psi)-h(\varphi)$, the sum over a closed sequence telescopes to $h(\varphi_0)-h(\varphi_0)=0$. Conversely, let $F$ be loop-neutral and set $h(x)=F(0,x)$. The loop $0\to0$ gives $F(0,0)=0$. The loop $0\to x\to0$ gives $F(x,0)=-h(x)$. For any $\varphi,\psi$, the loop $0\to\varphi\to\psi\to0$ gives $h(\varphi)+F(\varphi,\psi)-h(\psi)=0$.
\end{proof}
Eq.~\ref{eq:trajectory-return} takes $h(x)=x$, the choice under which Theorem~\ref{thm:maximizer} holds without further calibration.

\begin{suppcor}[Cumulative reward]
\label{cor:cumulative}
The cumulative reward $F(\varphi,\psi)=\psi$ is not loop-neutral. Moreover, if $\alpha<\tfrac12$, there exist error-free trajectories $\tau_{\mathrm f}$ with $d_T=0$ and $\tau_{\mathrm s}$ with $d_T=1$ such that $\sum_t\widetilde{\Phi}_t(\tau_{\mathrm f})>\sum_t\widetilde{\Phi}_t(\tau_{\mathrm s})$ for every $T>\tfrac{1}{1-2\alpha}$, while $R(\tau_{\mathrm s})=1>1-\alpha\geq R(\tau_{\mathrm f})$.
\end{suppcor}
\begin{proof}
The one-step loop $\varphi\to\varphi$ with $\varphi>0$ earns $\varphi\neq0$. For the second claim, let $\tau_{\mathrm f}$ verify all non-answer nodes by step $1$ and hold them until $T$ without answering, so $\widetilde{\Phi}_t=1-\alpha$ for all $t$. Let $\tau_{\mathrm s}$ progress at a constant rate, $\widetilde{\Phi}_t=t/T$, and complete at $T$. Then $\sum_t\widetilde{\Phi}_t(\tau_{\mathrm f})-\sum_t\widetilde{\Phi}_t(\tau_{\mathrm s})=T(1-\alpha)-\tfrac{T+1}{2}=\tfrac{T(1-2\alpha)-1}{2}$, which is positive under the stated condition on $T$. The returns follow from Theorem~\ref{thm:maximizer}.
\end{proof}
With unit weights, $\alpha=1/|V|<\tfrac12$ whenever $|V|\geq3$. The cumulative objective then prefers holding partial evidence without answering over steady progress to completion, and the preference grows linearly with the horizon. The incremental return ranks the same pair correctly at every horizon.

\subsection{Sign Consistency of the Action Weight}
\begin{suppprop}
\label{prop:sign}
For every action, $\operatorname{sign}(A_t)=\operatorname{sign}(r_t)$.
\end{suppprop}
\begin{proof}
If $r_t=0$, then $s_t=0$ and $g(0)=0$, so $A_t=0$. If $r_t>0$, then $A_t=[u_t]_++g(r_t)\geq g(r_t)>0$. If $r_t<0$, then $A_t=-[-u_t]_++g(r_t)\leq g(r_t)<0$.
\end{proof}
Centered advantages lack this property. In a bin with rewards $-0.1$ and $-0.3$, the first action has $u_t=+1$ although its reward is negative. At $\rho=1$ the gradient contribution of a selected action is $A_t\nabla_\theta\log\pi_\theta(a_t\mid h_{t-1})$. By Proposition~\ref{prop:sign}, a step along this direction raises the probability of the action, to first order, exactly when $r_t>0$. Selecting actions by $|A_t|$ preserves the sign, so every retained update moves the policy toward verified progress and away from errors.

\subsection{Sensitivity to Verification Errors}
\label{sec:supp-progress-sensitivity}
The potential-difference objective permits a direct analysis of verification errors on a fixed recorded trajectory. Let $\widehat{\Phi}_t=\widetilde{\Phi}_t+\eta_t$ be a perturbed training potential, where $\eta_t$ includes the effect of local evidence errors and their propagation through the dependency graph. Compute each transition difference from these consistently recorded potentials. The change in progress return is
\begin{equation}
 \begin{aligned}
 \widehat{R}_{\mathrm{prog}}-R_{\mathrm{prog}}
 &=\sum_{t=1}^{T}(\eta_t-\eta_{t-1})\\
 &=\eta_T-\eta_0.
 \end{aligned}
 \label{eq:supp-progress-error}
\end{equation}
Thus, if $|\eta_t|\leq\varepsilon_{\Phi}$, the absolute progress-return error is at most $2\varepsilon_{\Phi}$, or $\varepsilon_{\Phi}$ when the initial potential is fixed exactly. By comparison, a cumulative state-potential return changes by $\sum_{t=1}^{T}\eta_t$, whose worst-case magnitude is $T\varepsilon_{\Phi}$. Differencing keeps a persistent potential error from accumulating with trajectory length.

Let $\widehat{\ell}_t=\ell_t+\xi_t$ denote the perturbed execution cost, with $\ell_t=e_t/|V|$ in Eq.~\ref{eq:trajectory-return}. The resulting changes in local reward and episode return are
\begin{equation}
 \begin{aligned}
 \widehat r_t-r_t &= \eta_t-\eta_{t-1}-\xi_t,\\
 \widehat R-R &= \eta_T-\eta_0-\sum_{t=1}^T\xi_t .
 \end{aligned}
 \label{eq:supp-total-error}
\end{equation}
An error confined to an intermediate potential can change the weights of adjacent actions even when its contributions to the episode return cancel. The episode-sum bound applies to fixed traces. Action weights are computed from the perturbed transition rewards and their comparison bins, so their sensitivity also depends on reward signs and within-bin variation.

\section{Benchmark Card and Training Recipe}
\label{sec:supp-benchmark-training}

\paragraph{ViInteract construction.}
The ViInteract release contains 76,000 SFT episodes across nine source components, ten execution domains, and 34 task-program types; the 15,000-episode RL source pool and the 4,000-candidate benchmark pool each cover eight source components, nine execution domains, and 29 task-program types. The additional SFT component is COCO spatial relation, with five relation-program types. Evaluation panels are derived from the benchmark candidate pool. In Core200, the Memory Control component contributes 16 temporal-memory episodes and 2 selective-identity episodes, which is why the component-to-domain expansion contains nine domains. Core200 is an application-weighted panel with fixed component/length quotas. It contains 200 frozen episodes, 781 distinct source-image IDs, with no source-image ID shared between episodes; its payload manifest has SHA256 prefix \texttt{4b71dd74}. Holdout200 is selected after excluding Core200 and the earlier diagnostic panels, contains 787 unique source images, and has manifest SHA256 prefix \texttt{80343458}. Holdout200 is the held-out release used for transfer results. The two releases use the same public interaction protocol: only the current image is visible, the previous image is evicted after \texttt{NEXT\_IMAGE}, visual tools return typed observations, and memory records require an evidence reference to a successful tool event.

\paragraph{Task-program construction and validation.}
We build episodes from existing visual data using parameterized task-program templates. Source annotations supply reference observations; deterministic rules instantiate graph nodes, prerequisite edges, expected answers, and scripted interactions. LLM-assisted paraphrasing produces instruction wording from this task specification. A gold replay then executes the program under the public runtime contract to check that the constructed task can be completed. Every task program has the final-answer node as its unique sink. Each edge $j\to i$ is also a provenance requirement of the child predicate: $b_i$ holds only if node $i$'s evidence cites an event produced by node $j$. The graph edges therefore coincide with the evidence references checked by the verifier.

Table~\ref{tab:supp-benchmark-card} lists the execution domains, and Table~\ref{tab:supp-related-settings} compares ViInteract with related visual-agent settings.

\paragraph{Reconstructing the frozen panels.}
Tables~\ref{tab:supp-panel-quotas} and~\ref{tab:supp-panel-composition} specify the frozen panel composition. The panel builder is deterministic given the source pool, exclusion manifests, and seed. For every candidate, we first assign its source component and turn-length bucket $\{1,2,3,4$--$5,6$--$8,9$--$12,13+\}$. For the six short-capable components (ChartQA, COCO Detection, COCO Pose, LVIS Segmentation, SOP Identity, and TextOCR), one-, two-, and three-turn controls are derived from the first qualifying visual turns; the derived task program keeps only nodes verified by a gold replay and rewrites the final turn to the public answer schema. Existing four-or-more-turn episodes are retained without shortening. Candidates whose source episode or image path occurs in an exclusion manifest are removed before selection. Within each component/length cell, rows are sorted by $\operatorname{SHA256}(s\mathbin{:}\!\!\operatorname{episode\_id})$ and greedily accepted until the predeclared quota is filled, rejecting repeated source IDs and repeated source images. The selected rows are sorted by the same rank, replayed through the official runtime, and frozen with a payload hash. The Core200 and Holdout200 builders use seed 20260907; Holdout200 applies the same procedure after excluding Core200 and all earlier diagnostic panels. No model prediction is read during selection.

\begin{table*}[t]
\centering\scriptsize\setlength{\tabcolsep}{3.2pt}\renewcommand{\arraystretch}{1.12}
\caption{\textbf{Exact panel quotas used by the deterministic builder.} A dash denotes an empty component/length cell. The component totals reproduce the Core200 and Holdout200 panels; the resulting manifests and source-pool hash identify the released instances.}
\label{tab:supp-panel-quotas}
\begin{tabular*}{\textwidth}{@{\extracolsep{\fill}}lrrrrrrr|rrrrrrr@{}}
\toprule
 & \multicolumn{7}{c|}{Core200} & \multicolumn{7}{c}{Holdout200} \\
Source component & 1 & 2 & 3 & 4--5 & 6--8 & 9--12 & 13+ & 1 & 2 & 3 & 4--5 & 6--8 & 9--12 & 13+ \\
\midrule
COCO Detection & 5 & 15 & 15 & 15 & -- & -- & -- & 5 & 14 & 14 & -- & -- & -- & -- \\
TextOCR & 2 & 8 & 12 & 6 & 7 & -- & -- & 2 & 8 & 12 & 6 & 3 & -- & -- \\
ChartQA & 2 & 10 & 8 & 10 & -- & -- & -- & 2 & 10 & 8 & 3 & -- & -- & -- \\
COCO Pose & 1 & 3 & 3 & 4 & 9 & -- & -- & 1 & 3 & 3 & 4 & -- & -- & -- \\
LVIS Segmentation & -- & 3 & 2 & 2 & 8 & -- & -- & -- & 3 & 2 & 2 & 8 & -- & -- \\
SOP Identity & -- & 3 & 2 & 3 & 7 & -- & -- & -- & 3 & 2 & 3 & 7 & -- & -- \\
AITW Temporal & -- & -- & -- & 3 & 3 & 5 & 6 & -- & -- & -- & 40 & 3 & 5 & 6 \\
Memory Control & -- & -- & -- & 3 & 3 & 6 & 6 & -- & -- & -- & 3 & 3 & 6 & 6 \\
\midrule
Total & 10 & 42 & 42 & 46 & 37 & 11 & 12 & 10 & 41 & 41 & 61 & 24 & 11 & 12 \\
\bottomrule
\end{tabular*}
\end{table*}

\begin{table}[!tbp]
\centering\small
\setlength{\tabcolsep}{3.5pt}\renewcommand{\arraystretch}{1.1}
\caption{\textbf{ViInteract benchmark card.} Counts refer to the frozen Core200 composition.}
\label{tab:supp-benchmark-card}
\begin{tabularx}{\columnwidth}{@{}XrX@{}}
\toprule
Component / domain & $N$ & Primary execution requirement \\
\midrule
COCO Detection / visual-tool & 50 & locate and verify objects \\
TextOCR / OCR-memory & 35 & extract, write, retrieve text \\
ChartQA / multi-chart & 30 & compare chart evidence \\
COCO Pose / pose-memory & 20 & measure pose relations \\
Memory Control / temporal memory & 16 & preserve state across turns \\
Memory Control / selective identity & 2 & retain identity evidence \\
AITW Temporal / temporal UI & 17 & execute ordered actions \\
SOP Identity / multi-view identity & 15 & match across views \\
LVIS Segmentation / mask-measure & 15 & segment and measure \\
\midrule
\multicolumn{2}{r}{Total} & 200 episodes \\
\bottomrule
\end{tabularx}
\end{table}

\begin{table*}[t]
\centering\scriptsize\setlength{\tabcolsep}{4pt}\renewcommand{\arraystretch}{1.12}
\caption{\textbf{Positioning against active visual reasoning settings.} The table compares the execution contract and learning target. A checkmark denotes an explicit part of the reported setting; a dash denotes that the capability is outside the benchmark or method target.}
\label{tab:supp-related-settings}
\begin{tabularx}{\textwidth}{@{}lXXXXXX@{}}
\toprule
Setting & Multi-turn visual input & Active visual operation & Typed tools & Persistent evidence & Dependency-gated success & Primary learning target \\
\midrule
ViGoRL~\cite{sarch2025vigorl} & \checkmark & Grounding / zoom & -- & -- & -- & Spatially grounded reasoning \\
MGPO~\cite{huang2025mgpo} & \checkmark & Multi-turn crop & -- & -- & -- & High-resolution visual reasoning \\
DeepEyes~\cite{zheng2025deepeyes} & \checkmark & Image operations & -- & -- & -- & Thinking with images \\
ACTIVE-o3~\cite{zhu2025activeo3} & \checkmark & Active perception & Modular sensing & -- & -- & Information acquisition \\
MuirBench~\cite{wang2024muirbench} & Multi-image & -- & -- & -- & -- & Multi-image understanding \\
MMLongBench-Doc~\cite{ma2024mmlongbenchdoc} & Long visual document & -- & -- & -- & -- & Long-context document understanding \\
\rowcolor{resultshade}
\textbf{ViInteract + ViStep-RL} & \textbf{Streaming images} & \textbf{Learned tool policy} & \textbf{\checkmark} & \textbf{\checkmark} & \textbf{\checkmark} & \textbf{Evidence-state progress} \\
\bottomrule
\end{tabularx}
\end{table*}

\begin{table}[!tbp]
\centering\small
\setlength{\tabcolsep}{3.5pt}\renewcommand{\arraystretch}{1.1}
\caption{\textbf{Panel composition by source component and episode length.} Counts are frozen before the corresponding evaluation matrix.}
\label{tab:supp-panel-composition}
\begin{tabularx}{\columnwidth}{@{}Xrr@{}}
\toprule
Source component & Core200 & Holdout200 \\
\midrule
AITW temporal & 17 & 54 \\
ChartQA & 30 & 23 \\
COCO detection & 50 & 33 \\
COCO pose & 20 & 11 \\
LVIS segmentation & 15 & 15 \\
Memory control & 18 & 18 \\
SOP identity & 15 & 15 \\
TextOCR & 35 & 31 \\
\midrule
Length 1 & 10 & 10 \\
Length 2 & 42 & 41 \\
Length 3 & 42 & 41 \\
Length 4--5 & 46 & 61 \\
Length 6--8 & 37 & 24 \\
Length 9--12 & 11 & 11 \\
Length 13+ & 12 & 12 \\
\bottomrule
\end{tabularx}
\end{table}

Table~\ref{tab:supp-release-stats} reports the release size and structural statistics.
\paragraph{Data isolation audit.}
The SFT snapshot contains 76,000 training episodes and the GRPO snapshot contains 15,000 training episodes. We compared each training snapshot with Core200, Holdout200, and the 500-item process panel using episode IDs, complete image paths, and source-sequence IDs. Every comparison has zero overlap; Core200 and Holdout200 are also disjoint on all three keys. The raw numeric image ID is a local identifier reused across source components and is therefore not used as a leakage key. The 500 episodes used for the reward comparisons in main-text Table~\ref{tab:reward-ablations} have payload hash prefix \texttt{00930e46}. A separate check of these episodes, their 2,500 distinct image paths and 215 source sequence identifiers found zero overlap with either the SFT or RL training split on these keys. The composition tables give the component and length mix of each panel; Holdout200 contains 54 AITW temporal episodes and Core200 contains 17.

\begin{table}[!tbp]
\centering\scriptsize
\setlength{\tabcolsep}{2pt}\renewcommand{\arraystretch}{1.12}
\caption{\textbf{Frozen release statistics.} Core200 and Holdout200 are disjoint evaluation releases; prefixes identify the payload manifests recorded in the release ledger.}
\label{tab:supp-release-stats}
\begin{tabular}{@{}lrrrrr@{}}
\toprule
Release & Images/ep. & Turns/ep. & Nodes/ep. & Unique images & ID \\
\midrule
Core200 & 1--16 & 1--16 & 3--33 & 781 & 4b71 \\
Holdout200 & 1--16 & 1--16 & 3--33 & 787 & 8034 \\
\midrule
Mean (Core) & 4.03 & 4.82 & 10.80 & -- & -- \\
Mean (Holdout) & 4.05 & 4.66 & 10.12 & -- & -- \\
\bottomrule
\end{tabular}
\end{table}

\paragraph{Episode semantics.}
An episode is a sequence of public turns with a current image, a task instruction, legal actions, and returned tool or memory observations. The global action budget is 32 for the reported final-answer protocol; generation is capped at 768 tokens and image pixels at 262,144. The hidden task program defines dependency-gated nodes for evidence acquisition, cross-image reasoning, memory writing, memory retrieval, and the final answer. \emph{Answer} checks the final structured response. \emph{Strict} additionally requires every dependency-gated node to be verified with the required provenance. The policy never receives the hidden node identifiers or expected evidence values.

\paragraph{Graph construction rules.}
\label{sec:supp-graph-construction}
The compiler builds each graph from the evidence that the answer schema cites. An edge $j\to i$ is created exactly when the predicate of node $i$ cites an event produced by node $j$, such as a memory write citing the OCR call that produced its payload; no other edges are added. The answer node is appended last and receives an edge from every node that has no other child, so it is the unique sink and every node is an ancestor of it. The answer node carries no retrieval coverage. Positive weights default to one. These rules give properties (A1)--(A4) of Section~\ref{sec:supp-theory} for every released program.

\paragraph{Task-graph authoring and verification.}
Each released episode stores a task program whose nodes are derived from source annotations, typed tool outputs, deterministic computations, memory operations, and the final answer schema. A node records its type, expected value or source annotation, parent node identifiers, and an optional weight; the runtime assigns weight one when no weight is specified. The verifier accepts a visual node only when the corresponding typed observation contains the expected evidence. A memory node additionally requires a matching payload or claim, an image or semantic identity, and evidence references that resolve to successful evidence-producing calls in the same trajectory. A retrieval node is scored during training by $|\mathcal{R}_i\cap\mathcal{M}_i|/|\mathcal{M}_i|$, but is strict-verified only at complete coverage. The verifier recomputes the dependency fixed point after every transition, so deleting or corrupting a memory record retracts that node and every dependent node. The same fixed point defines both the Strict metric and the progress signal.

\paragraph{Top-level release schema and verifier fixed point.}
Tables~\ref{tab:supp-json-schema} and~\ref{tab:supp-task-schema} distinguish public record fields from hidden verifier fields.
Each frozen JSON object has an immutable public payload and a hidden evaluation payload. The public fields define what the policy can observe; the hidden fields define what the scorer can verify. After every action, the runtime (1) applies the public state transition, (2) records the typed observation and action provenance, (3) checks evidence and memory payloads, (4) activates nodes whose local predicates and parents are satisfied, and (5) repeats activation until the node set reaches a fixed point. It then emits the verified node set, retrieval coverage, and the before/after potentials. A state update that invalidates a parent automatically retracts all dependent nodes before the next credit difference is computed.

\begin{table}[!tbp]
\centering\scriptsize\setlength{\tabcolsep}{3pt}\renewcommand{\arraystretch}{1.08}
\caption{\textbf{Top-level ViInteract release schema.} Public fields are visible to the policy; hidden fields are consumed only by the verifier and scorer.}
\label{tab:supp-json-schema}
\begin{tabularx}{\columnwidth}{@{}lXl@{}}
\toprule
Field & Stored content & Visibility \\
\midrule
\texttt{id}, \texttt{episode\_id} & Stable episode and source identifiers & Public \\
\texttt{dataset\_version}, \texttt{partition}, \texttt{split} & Release and partition metadata & Public \\
\texttt{domain}, \texttt{release\_component} & Task family and source component & Public \\
\texttt{source\_sequence\_id}, \texttt{images} & Source sequence and ordered image records & Public \\
\texttt{student\_visible} & Instruction, current image, legal actions, observations & Public \\
\texttt{tool\_state} & Typed tool outputs and action provenance & Public \\
\texttt{task\_program} & Nodes, dependencies, expected evidence, weights & Hidden \\
\texttt{hidden\_eval} & Strict predicates and answer checks & Hidden \\
\texttt{counterfactual\_protocol} & Intervention hooks for memory/tool studies & Hidden \\
\texttt{metadata}, \texttt{immutable\_payload\_hash} & Construction metadata and payload identity & Hidden \\
\bottomrule
\end{tabularx}
\end{table}

\begin{table}[!tbp]
\centering\scriptsize\setlength{\tabcolsep}{3pt}\renewcommand{\arraystretch}{1.12}
\caption{\textbf{Task-program fields and verification rules.} Hidden fields are consumed by the runtime and scorer; they are excluded from the policy observation.}
\label{tab:supp-task-schema}
\begin{tabularx}{\columnwidth}{@{}lXl@{}}
\toprule
Field & Stored content & Runtime use \\
\midrule
\texttt{node\_id} & Stable node identifier & Dependency references \\
\texttt{type} & Evidence, memory, retrieval, computation, or answer & Selects verifier \\
\texttt{depends\_on} & Parent node identifiers & Fixed-point ordering \\
\texttt{expected} & Source value, annotation, or deterministic predicate & Evidence match \\
\texttt{evidence\_refs} & Producing tool-call identifiers & Provenance check \\
\texttt{weight} & Optional positive node weight & Potential normalization \\
\texttt{memory\_refs} & Required memory identifiers/fields & Write and retrieval checks \\
\bottomrule
\end{tabularx}
\end{table}

\begin{table}[!tbp]
\centering\small
\setlength{\tabcolsep}{3pt}\renewcommand{\arraystretch}{1.12}
\caption{\textbf{Qwen policy update.} The verifier evaluates every environment transition; each selected action receives its own signed weight.}
\label{tab:supp-training-path}
\begin{tabular}{@{}p{0.08\columnwidth}p{0.87\columnwidth}@{}}
\toprule
Step & Operation and emitted state \\
\midrule
1 & Sample a rollout group under the current policy. \\
2 & Execute each action and record the public observation, typed tool result, memory provenance, and verifier state. \\
3 & Recompute the dependency fixed point and evaluate $\delta_t$ and the execution cost. \\
4 & Check variation in rollout returns, then compare transition rewards within phase, turn, and progress bins. \\
5 & Select actions by absolute weight and apply the clipped token loss with the frozen SFT reference. \\
\bottomrule
\end{tabular}
\end{table}

\paragraph{Controlled evaluation interfaces.}
The memory factorial uses the frozen Core200 release and the same 3B trained weights as the native comparison. All four conditions run under a wrapper with at most eight stored records, 512 serialized characters per record and four records per retrieval. Active writing lets the policy select what to store, while automatic writing stores observations through the wrapper. Storage limits and retrieval limits are shared. Actual write counts and occupied storage depend on the policy. The Active/Active row is evaluated with these constraints.

The model--Harness factorial uses the same Core200 episode IDs in every condition. Its Direct arm exposes images sequentially and accumulates public observation notes for a common final-answer request. The Harness arm enables tool calls and manages image progression, context and memory. The native training-stage comparison instead asks the policy to act through the agent protocol, including legal tool calls and answer submission. Answer scores are compared within each interface.

The Standard GRPO baselines use the same SFT initialization, tools, action budget and training volume as ViStep-RL at each scale. The baseline rewards terminal-answer correctness and normalizes episode returns within each rollout group, sharing the resulting advantage across the trajectory. In the 7B terminal-reward control with action-level updates, only the final answer action receives the answer reward. All preceding actions receive zero reward and therefore zero signed reward weight. Answer and Strict are reported on Core200 and the remaining four metrics on Process500.

\paragraph{Action weights and policy loss.}
For each prompt we sample eight trajectories and admit the group when the population standard deviation of its returns exceeds $10^{-5}$. All groups in this comparison share the same prompt. Any other group contributes zero policy-loss weight. A comparison bin $B$ contains actions with the same phase, visual turn, and $\operatorname{round}(10\widetilde{\Phi}_{t-1})$. For bin mean $\mu_B$ and population standard deviation $\sigma_B$ of transition rewards, we use $u_t=(r_t-\mu_B)/(\sigma_B+10^{-4})$, with $u_t=0$ when $\sigma_B\leq10^{-6}$. In Eq.~\ref{eq:group-advantage}, $g(r)=0.25\,\operatorname{clip}(r/0.05,-1,1)$. Rewards with $|r_t|\leq10^{-8}$ receive zero weight. A singleton bin or a bin with identical rewards has zero standardized term. Its nonzero transition rewards still contribute through $g$. A zero transition reward always receives zero weight.

For a generated token $y$ with prefix $h$ in selected action $t$, let $\rho=\pi_\theta(y\mid h)/\pi_{\rm old}(y\mid h)$, where $\pi_{\rm old}$ sampled the action. Let $v=\log\pi_{\rm ref}(y\mid h)-\log\pi_\theta(y\mid h)$ for the frozen SFT reference. The per-token loss is
\begin{equation}
 \begin{aligned}
 L={}&-\min\{\rho A_t,\operatorname{clip}(\rho,0.8,1.2)A_t\}\\
 &+0.03\,(\exp(v)-v-1).
 \end{aligned}
 \label{eq:supp-policy-objective}
\end{equation}
We average first over generated tokens within an action, then over selected actions within a rollout, and finally over rollouts. The loss uses immediate transition weights from main-text Eq.~\ref{eq:group-advantage}. The episode return is used for group admission. Selected actions are ranked by absolute weight, with seeded random ordering for ties. We select up to two actions per resumed 3B rollout and four per 7B rollout. When fewer actions are available, padding entries receive zero loss weight.

\paragraph{Training recipe.}
The primary Qwen runs use rank-16 LoRA with alpha 32, learning rate $2\times10^{-7}$, seed 42, eight rollouts per prompt, temperature 0.9, and top-$p$ 0.95. Training uses bfloat16 precision, a 768-token generation cap and at most 262,144 pixels per image. The 7B run caps prompts at 8,192 tokens and rollouts at 32 actions, selecting up to four actions per rollout. The resumed 3B run caps prompts at 11,520 tokens and rollouts at 16 actions, selecting up to two. For each transition, the environment records the public state, action, observation, memory provenance, and the before/after training potential. The primary Qwen2.5-VL-7B run completed 1,000 updates and 32,000 sampled rollouts on four GPUs, using the final saved adapter for evaluation. The primary native Qwen2.5-VL-3B policy is the step-700 checkpoint; the matched Holdout200 write-disable and corruption runs use the same 3B policy as their intact control.

\paragraph{Credit computation.}
Table~\ref{tab:supp-training-path} describes how the recorded Qwen transitions enter the policy update in main-text Eq.~\ref{eq:group-advantage} and Eq.~\ref{eq:supp-policy-objective} above.

\paragraph{External evaluation protocol.}
On MMLongBench-Doc, MuirBench and MMSI-Bench, we compare Qwen2.5-VL-3B ViStep-RL with a reference model of the same backbone. Within each benchmark, both models receive identical inputs and use the same evaluation prompt.

\section{Execution Analysis on Core200}
Table~\ref{tab:supp-efficiency} reports the interaction cost of Qwen2.5-VL-7B on the same Core200 episodes. ViStep-RL completes more episodes than SFT with fewer actions, fewer action tokens, and less time per episode.
\begin{table}[H]\centering\small\setlength{\tabcolsep}{3pt}\renewcommand{\arraystretch}{1.15}
\caption{\textbf{Interaction efficiency on Core200.} Qwen2.5-VL-7B; values are per episode.}
\label{tab:supp-efficiency}
\begin{tabularx}{\columnwidth}{@{}l*{3}{>{\raggedleft\arraybackslash}X}@{}}
\toprule
Stage & Actions & Action tokens & Time (s) \\
\midrule
Base & 19.54 & 1,386.8 & 44.87 \\
SFT & 15.76 & 1,921.0 & 72.46 \\
\rowcolor{resultshade}
\textbf{ViStep-RL} & \textbf{15.00} & 1,732.2 & 67.43 \\
\bottomrule
\end{tabularx}
\end{table}
The paired ViStep-RL minus SFT action-count difference is $-0.765$ actions per episode. Action tokens are estimated by re-tokenizing saved action text; time is the per-episode share of batch wall time.

The native 7B comparison uses 200 matched Core200 episodes. ViStep-RL gains 32 Strict successes over SFT and loses five, a net gain of 27 episodes (13.5 points). The 95\% paired percentile bootstrap interval is [8.0, 19.0] points over 10,000 episode-ID resamples with seed 42, and the two-sided exact McNemar test gives $p=7.43\times10^{-6}$.

\begin{figure*}[ht]
\centering
\includegraphics[width=0.96\textwidth]{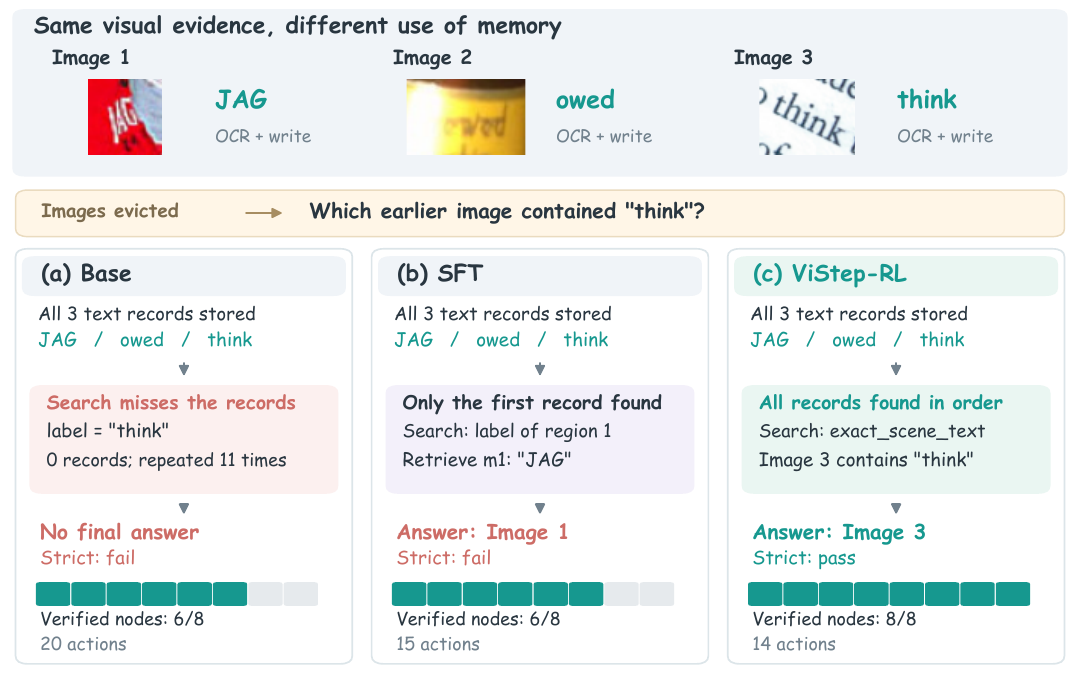}
\caption{\textbf{Same evidence, different task completion.} Matched Qwen2.5-VL-7B trajectories on one TextOCR episode. All stages read and store the same evidence; after image eviction, only ViStep-RL retrieves all required records and answers correctly. Counts cover the full recorded trajectories.}
\label{fig:trajectory-comparison}
\end{figure*}

\section{Human Evaluation Protocol}
\label{sec:supp-human}
\paragraph{Sample and raters.}
The study covers all 200 Core200 episodes and the Qwen2.5-VL-7B traces of SFT, Standard GRPO and ViStep-RL, giving 600 traces. Three annotators rate every trace independently. Model names are hidden, and the three traces of an episode appear in random order.

\paragraph{Rating instructions.}
Annotators first see the task instruction, the source images, the tool outputs and the memory write and retrieval log, with the final answer hidden. They answer three binary questions and one graded question.
\begin{itemize}[leftmargin=*,itemsep=1pt,topsep=2pt]
\item \emph{Evidence validity}: every stored record matches the content of its source image.
\item \emph{Provenance validity}: every stored record names its source image and the tool event that produced it.
\item \emph{Downstream support}: the records retrieved before the final decision suffice to determine the answer.
\item \emph{Helpfulness}: how much the trace helps a reader check the answer, from 1 (not at all) to 5 (fully).
\end{itemize}
A binary criterion holds for a trace when at least two annotators accept it. A trace is \emph{rated helpful} when at least two annotators give a score of 4 or higher. Intervals are percentile intervals from 10,000 episode-level bootstrap resamples with seed 0.

\paragraph{Agreement.}
\begin{table}[H]\centering\small
\caption{\textbf{Inter-annotator agreement} over all 600 traces.}
\label{tab:supp-human-agreement}
\begin{tabular}{@{}lrr@{}}\toprule
Criterion & Fleiss' $\kappa$ & Unanimous (\%)\\\midrule
Evidence validity & 0.744 & 88.2\\
Provenance validity & 0.716 & 79.7\\
Downstream support & 0.674 & 75.7\\
\bottomrule\end{tabular}
\end{table}
All three binary criteria reach substantial agreement. For helpfulness, 89.8\% of annotator pairs give scores within one point of each other.

\section{Qualitative Trajectory Comparison}
Figure~\ref{fig:trajectory-comparison} follows the same TextOCR episode across all three training stages. All stages successfully read and store ``JAG,'' ``owed,'' and ``think'' from images 1--3, respectively. The three stages therefore differ only in how they use the stored evidence. The final question is asked with no image visible. In each node strip, the first six blocks represent three OCR and three memory-write nodes; the last two represent task-level retrieval and final-answer verification. Teal denotes verified nodes and gray denotes unverified nodes.

Base searches for a record whose label is ``think,'' although its stored labels describe source regions, and receives no records in 11 consecutive searches. SFT searches using the first region's label and retrieves only the ``JAG'' record, then answers Image 1. ViStep-RL searches by the shared claim \texttt{exact\_scene\_text}, obtains all three records ordered by image index, and answers Image 3. Final task verification is 6/8, 6/8, and 8/8 nodes for Base, SFT, and ViStep-RL, respectively, with 20, 15, and 14 total actions. Completing the episode requires retrieving all three records with a query that matches how they were stored.

\section{General-Purpose Model Evaluation}
\label{sec:api-supp}
\paragraph{Evaluation protocol.}
Harness runs use the 200 Core200 episode IDs and Direct runs use the matched 200-episode Direct panel of the main paper. All API runs use temperature 0, seed 42, one final submission, a 32-action global tool budget, at most four tools per image, and at most eight tools in the final phase. Generation is capped at 768 tokens and images at 262,144 pixels. Direct supplies rolling image-observation notes; Harness adds tool dispatch and retains the returned public tool observations. Every endpoint receives the same public instructions, and ViStep-RL runs inside the same Harness. Table~\ref{tab:supp-api-paired} gives paired differences on the shared Harness.

\begin{table}[H]\centering\small
\caption{Paired advantage of the trained Qwen2.5-VL-7B with Harness over each API model with Harness. Differences are percentage points on the same 200 items.}
\label{tab:supp-api-paired}
\begin{tabular}{lrr}\toprule
API model & Difference & 95\% CI \\\midrule
GPT-4o & +24.5 & [17.5, 31.5] \\
Claude Opus 4.6 & +29.0 & [21.0, 37.0] \\
Qwen3-VL-Plus & +28.0 & [20.0, 36.0] \\
Qwen3.8-Flash & +24.0 & [16.5, 31.5] \\
Gemini 3.8 Flash & +25.0 & [17.0, 32.5] \\
Kimi K2.5 & +37.5 & [29.5, 45.5] \\
\bottomrule\end{tabular}\end{table}
Intervals use 10,000 paired episode-ID bootstrap resamples with seed 42 and lexicographically sorted IDs. Each resample keeps both models' outcomes for an episode together. These intervals describe uncertainty across the 200 evaluated episodes.

\end{document}